\documentclass{article}

\usepackage{microtype}
\usepackage{graphicx}
\usepackage{subfigure}
\usepackage{booktabs} 
\usepackage{amsmath}
\usepackage{amsfonts}
\usepackage{amssymb}
\usepackage[frozencache=true,cachedir=minted-cache]{minted}

\usepackage{hyperref}

\usepackage{pifont}
\usepackage{xcolor}
\newcommand{\cmark}{\textcolor{green!80!black}{\ding{51}}}
\newcommand{\xmark}{\textcolor{red}{\ding{55}}}

\DeclareMathOperator*{\argmax}{argmax}

\newsavebox{\mintedbox}
\usepackage{tcolorbox}
\tcbuselibrary{minted,breakable,skins}
\usepackage{listings}
\usepackage{caption}
\newenvironment{code}{\captionsetup{type=listing}}{}

\usepackage[accepted]{mlsys2025}

\mlsystitlerunning{PyRelationAL}

\begin{document}

\twocolumn[
\mlsystitle{PyRelationAL: A framework and toolkit for active learning research}




\begin{mlsysauthorlist}
\mlsysauthor{Paul Scherer}{rrx}
\mlsysauthor{Alison Pouplin}{rrx,goo}
\mlsysauthor{Alice Del Vecchio}{rrx}
\mlsysauthor{Suraj M.S.}{rrx}
\mlsysauthor{Oliver Bolton}{rrx}
\mlsysauthor{Jyothish Soman}{rrx}
\mlsysauthor{Jake P. Taylor-King}{rrx}
\mlsysauthor{Lindsay Edwards}{rrx}
\mlsysauthor{Thomas Gaudelet}{rrx}
\end{mlsysauthorlist}

\mlsysaffiliation{rrx}{Relation Therapeutics, Regent’s Place, 338 Euston Road, London NW1 3BG}
\mlsysaffiliation{goo}{Aalto University, Finland}

\mlsyscorrespondingauthor{Paul Scherer}{paul.scherer@relationrx.com}
\mlsyscorrespondingauthor{Thomas Gaudelet}{thomas@relationrx.com}

\mlsyskeywords{Machine Learning, MLSys}

\vskip 0.3in

\begin{abstract}
\textit{Active learning} (AL) is a sub-field of ML focused on the development of methods to iteratively and economically acquire data by strategically querying new data points that are the most useful for a particular task. Here, we introduce PyRelationAL, an open source library for AL research. We describe a modular toolkit based around a two step design methodology for composing pool-based active learning strategies applicable to both single-acquisition and batch-acquisition strategies. This framework allows for the mathematical and practical specification of a broad number of existing and novel strategies under a consistent programming model and abstraction. Furthermore, we incorporate datasets and active learning tasks applicable to them to simplify comparative evaluation and benchmarking, along with an initial group of benchmarks across datasets included in this library. The toolkit is compatible with existing ML frameworks. PyRelationAL is maintained using modern software engineering practices --- with an inclusive contributor code of conduct --- to promote long term library quality and utilisation. 
PyRelationAL is available under a permissive Apache licence on \texttt{PyPi} and at \url{https://github.com/RelationRx/pyrelational}.

\end{abstract}
]



\printAffiliationsAndNotice{}  

\section{Introduction}
\label{submission}
In recent years, machine learning (ML) models, particularly those leveraging deep learning, have become increasingly important in both scientific research and broader applications. Their success is largely attributed to strong performance across a wide range of tasks and domains. However, this success has often relied on the availability of large, annotated datasets and well-designed inductive biases. Unfortunately, many crucial fields such as biology, chemistry, and medicine frequently operate in low data regimes \cite{gaudelet_drugs}, where gathering enough labelled data to train accurate models is prohibitively expensive or time-consuming \cite{smith2018less,abdelwahab2019active,hoi2006batch}. For instance, determining the synergistic effects of FDA-approved drugs would require millions of experiments, making such labelling efforts experimentally infeasible.

To address these challenges, Active Learning (AL) has emerged as a promising approach \cite{fedorov1972theory, burrsettlesbook}. AL aims to develop strategies that maximize model performance while minimizing the amount of labelled data required. This is achieved by actively selecting the most informative samples for labelling, allowing the model to learn more efficiently under data constraints. AL is a well-established concept, closely related to query learning in computational learning theory \cite{burrsettlesbook} and sharing connections with Bayesian optimization \cite{brochu2010tutorial} and reinforcement learning \cite{sutton2018reinforcement}. These fields aim to strategically explore an input space to optimize a performance metric, often under uncertainty. Recently, AL has seen a resurgence in interest, particularly in combination with deep learning and real-world applications, such as in medicine, where AL has been applied to cost-sensitive data labelling and model refinement \cite{ren2021survey, zhanreview, smith2018less, abdelwahab2019active, hie2020leveraging, bertin2023recover, mehrjou_genedisco_2021}. 

\begin{figure*}
\begin{center}
    
    \includegraphics[width=0.9\textwidth]{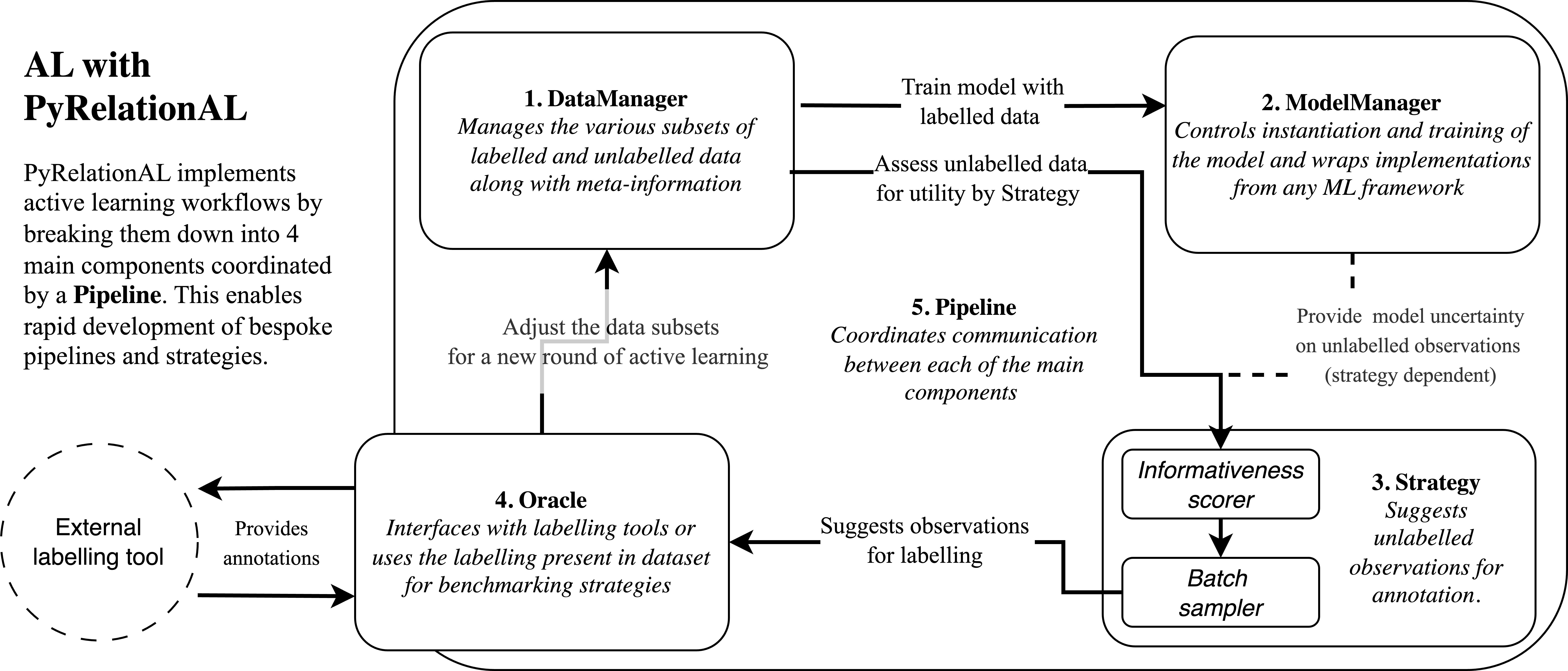}
\end{center}
    \caption{Diagram of PyrelationAL's modular approach to constructing full active learning pipelines.}
    \label{fig: framework}
\end{figure*}

Benchmarks, datasets, and software libraries have played a crucial role in driving progress in ML research. However, while other areas of ML have benefited from widely available benchmarks and standardized datasets, active learning suffers from fragmentation, with varying implementations and few shared resources for evaluation, especially large resources with enough labelled data for real-world tasks. This gap limits the reproducibility and scalability of research in active learning, making it harder for researchers and practitioners to compare and advance techniques consistently.

Here, we introduce PyRelationAL, an open-source Python library designed to facilitate the rapid, reliable, and reproducible construction of active learning pipelines. PyRelationAL adopts a modular design that simplifies the development of AL strategies, and the infrastructure around AL, and enables users to experiment with different components, from query selection methods to model uncertainty estimates, within a consistent framework. The library provides a wide range of pre-implemented AL strategies and makes it easy to design new ones or adapt existing ones to specific use cases, easing the significant engineering overhead required to implement AL systems. Furthermore, PyRelationAL integrates a number of datasets of varying size as part of its package as well as benchmarking functionalities proposing a way forward for active learning development and reducing the barrier of entry for both researchers and practitioners.

We summarize PyRelationAL's main contributions as follows:

\begin{itemize}
    \item \textbf{Modular toolkit:} Rapidly construct AL pipelines and strategies with reusable components. Particularly, decomposing AL strategies according to the two-step framework introduced in Section \ref{ssec:strat_design} facilitates modularity and flexible definition of AL strategies.
    \item \textbf{Pre-implemented strategies:} A comprehensive set of AL strategies allows for easy reproduction, modification, and extension.
    \item \textbf{Benchmark datasets:} Collection of synthetic and real-world datasets along with AL task configurations for evaluation across domains. 
    \item \textbf{Benchmarking:} Define benchmarks and collect results on a centralised platform  accelerating the AL development cycle. 
\end{itemize}

\section{Design and implementation}

PyRelationAL is built around five core modular components, designed to facilitate the rapid development of AL pipelines (Figure~\ref{fig: framework}). These pipelines have to encompass the infrastructure needed to facilitate interfacing between data management systems, annotators, ML models, AL strategies, logging, and more in a robust but also easy to use manner. These components are:

\begin{itemize}
    \item \textbf{DataManager}: The \texttt{DataManager} manages the labelled and unlabelled data pools within the AL workflow. It wraps around a PyTorch \texttt{Dataset}, tracks the split between labelled and unlabelled samples, and handles data loading through PyTorch's batch loader. This component ensures that the model has access to labelled data for training and can query from the unlabelled pool during each active learning iteration. 
    
    \item \textbf{ModelManager}: The \texttt{ModelManager} is responsible for interacting with the machine learning models. It handles model training, evaluation, and uncertainty estimation. PyRelationAL also provides implementations for uncertainty-aware models such as \texttt{MCDropoutModelManager} and \texttt{EnsembleModelManager}, allowing estimation of uncertainty under a point estimate model class or committee as part of their AL strategy. While the framework primarily supports PyTorch, the \texttt{ModelManager} can be adapted to any machine learning library. Note that the collate function used by the \texttt{DataManager} should also be adapted accordingly.
    
    \item \textbf{Active Learning Strategy}: The \texttt{Strategy} defines how the samples for labelling are selected and underpins the core research element of the field of active learning (see Appendix \ref{apppendix:primer}). PyRelationAL includes various built-in strategies like least confidence and margin sampling, which are based on compositions of different informativeness measures and query selection algorithms. We discuss PyRelationAL's two-step design philosophy for active learning strategies in more detail within Section \ref{ssec:strat_design}.
    
    \item \textbf{Oracle}: The \texttt{Oracle} component handles the interaction between the active learning strategy and the labelling process. For simulation purposes, the \texttt{BenchmarkOracle} allows users to perform AL experiments on pre-labelled datasets by simulating queries to an annotator. This can be modified to provide random and heteroscedastically influenced labels as well. For real-world applications, oracles can interface with actual labelling systems or external tools, providing flexibility across experimental and practical scenarios when adapted to the API.
    
    \item \textbf{Pipeline}: The \texttt{Pipeline} is the component that brings together the \texttt{DataManager}, \texttt{ModelManager}, strategy, and \texttt{Oracle} into a cohesive active learning loop. It manages the flow of data, executes each AL round, and updates the model with newly labelled samples. The pipeline design makes it straightforward to iterate over AL cycles, ensuring smooth integration between querying and model updating processes.
\end{itemize}

Section \ref{sec:case_study} provides a complete example on how to combine these components to design and run an active learning experiments.

\subsection{A Modular Framework for Pool-Based Active Learning} \label{ssec:strat_design}

We propose a modular, two-step approach to pool-based active learning, comprising:

\begin{enumerate} \item Informativeness Scoring: Estimating the potential benefit of labelling each sample, based on how much information it would contribute to model improvement. \item Batch Selection: Choosing an informative subset of samples for labelling, essential in batch-mode active learning. \end{enumerate}

This framework enables flexible combinations of scoring and sampling methods, providing a standardized way to experiment with various active learning strategies. Many popular methods can be reformulated to fit within this two-step structure, and new methods can seamlessly integrate into each component.


\begin{table}[ht]
    \caption{List of implemented informativeness scorers and batch samplers in PyRelationAL at time of writing. The listed names are the name of the classes in the repository. Appendix \ref{appendix:infos} provides additional details about informativeness measures.\vspace{0.1cm}}
    \label{tab:alstrategies}
    \begin{center}
\begin{tabular}{@{}ll@{}}
\toprule  
\multicolumn{2}{l}{\centering \textbf{Informativeness scorers}} \\ \midrule
\texttt{AverageScorer}\\
\texttt{ClassificationBALD} & \\
\texttt{Entropy} & \\
\texttt{ExpectedImprovement}& \\
\texttt{LeastConfidence} &  \\
\texttt{MarginConfidence} &  \\
\texttt{RatioConfidence} & \\
\texttt{RegressionBALD} & \\
\texttt{RelativeDistanceScorer} & \\
\texttt{StandardDeviation} & \\
\texttt{ThompsonSampling} & \\
\texttt{UpperConfidenceBound} & \\
\midrule
\multicolumn{2}{l}{\centering  \textbf{Batch-mode samplers}} \\ \midrule
\texttt{TopKSampler} & \\
\texttt{ProbabilisticSampler} & \\
\bottomrule
\end{tabular}
\end{center}
\end{table}

\subsubsection{Informativeness Scoring}

In the literature, sample informativeness is often associated with how uncertain a model is about what label the sample should have. Typically, the assumption is that samples with higher uncertainty are more informative as they help anchor the model in uncertain regions. Common strategies include least confidence, margin sampling, and BALD (Bayesian Active Learning by Disagreement) \cite{bald}, which assesses information gain based on posterior uncertainty. Methods like Query by Committee \cite{seung1992query} add a layer of model diversity, using disagreement among a set of models to gauge informativeness. 

Other strategies, like Expected Model Change \cite{cai2013maximizing}, score samples based on how much they would impact the model’s parameters. Additionally, representativeness-based approaches score each sample based on how representative of the current data distribution they are. For instance, density-weighted approaches \cite{cai2015active} assign a higher score to samples found in the denser region of the data distribution under the assumption that they are more representative of the data.

It is noteworthy that scoring methods borrowed from the reinforcement learning literature, such as Upper Confidence Bound \cite{de2021greed} and Thompson Sampling \cite{russo2018tutorial}, introduces a notion of exploration in the informativeness scoring itself. This can also be achieved through the batch selection step detailed in the next section.

Importantly, the commonality among all these approaches is that they first focus on quantifying the expected utility of a sample for improving the model, and thus fit neatly into PyRelationAL’s first step of computing informativeness scores. PyRelationAL includes a suite of pre-defined scorers in the \texttt{informativeness} module, covering uncertainty-based, representativeness-based, and other specialized metrics. Users can also create custom scorers by sub-classing the \texttt{AbstractScorer} class. Table \ref{tab:alstrategies} provides a list of the currently implemented informativeness scorers in PyRelationAL.

\subsubsection{Batch Selection}

Once informativeness scores are assigned, the batch selection step identifies a set of samples for labelling. Many active learning strategies rely on simple greedy selection, where the top $k$ samples are chosen based on informativeness scores. However, in batch-mode learning, diversity is often essential to avoid redundant samples and ensure broader data coverage. Strategies such as BatchBALD \cite{kirsch2019batchbald} enhance this approach by maximizing the joint information gain of the batch, so each selected sample provides unique value. Other methods, like diversity sampling \cite{monarch_human---loop_2021}, prioritize diversity in the batch, which helps cover more of the input space and increases model robustness. Approaches such as $\epsilon$-greedy also introduce randomness to encourage exploration and improve data representation in sampling.

PyRelationAL’s batch-mode samplers support both greedy and diverse selection methods. The framework allows users to implement batch selection strategies that prevent redundancies and ensure diversity, critical in applications requiring balanced exploration-exploitation trade-offs or handling noisy labels \citep{pmlr-v16-settles11a, kirsch2023stochastic}. Users can create custom scorers through sub-classing the abstract \texttt{BatchModeSampler} class. Table \ref{tab:alstrategies} provides a list of the currently implemented batch-mode samplers in PyRelationAL.

\subsubsection{Flexible, Composable Active Learning Strategies}

By separating active learning into informativeness scoring and batch selection, PyRelationAL supports a wide range of strategies through modular components. This approach builds on Settles’ foundational active learning framework \cite{burrsettlesbook} but extends it to accommodate both single-sample and batch selection methods within a unified structure. Users can readily combine scoring and sampling methods to create adaptive active learning strategies that align with specific research or application needs. Critically, the framework allows us to view existing strategies in a different way to consider new developments along either informativeness or batch selection steps.

Two core classes, \texttt{ClassificationStrategy} and \texttt{RegressionStrategy}, encapsulate these modular components, allowing users to customize strategies by specifying their preferred scorer and sampler.

While most active learning methods integrate smoothly into this framework, some methods (e.g., cluster-based representative sampling) operate outside of the two-step process, bypassing informativeness scoring by selecting cluster centers directly. Such methods are compatible with PyRelationAL through subclassing the abstract \texttt{Strategy} class. While these methods can be implemented, they may lack the reusability and flexibility of the core modular components.

\subsection{Use case study: Coding example}\label{sec:case_study}

In this section, we illustrate how to build an active learning pipeline using PyRelational's core components through a practical coding example. We focus on a digit classification task with the MNIST dataset, which is included in PyRelational (Table \ref{tab:aldatasets}).

We begin by defining a \texttt{DataManager} to handle the MNIST dataset. This involves specifying four different sets: labelled, unlabelled, validation, and test images.

\begin{code}

    \begin{minted}[linenos,fontsize=\scriptsize,xleftmargin=0.5cm,numbersep=3pt,frame=lines]{python}
from sklearn.model_selection import train_test_split

from pyrelational.data_managers import DataManager
from pyrelational.datasets.classification \
    import MNIST


dataset = MNIST()
train_ixs, test_ixs = dataset.data_splits[0]

unlabelled_ixs, val_ixs = train_test_split(
    train_ixs,
    test_size=0.1,
    random_state=42,
    stratify=dataset.y[train_ixs],
)

labelled_ixs, unlabelled_ixs = train_test_split(
    unlabelled_ixs,
    train_size=20,
    random_state=42,
    stratify=dataset.y[unlabelled_ixs],
)

data_manager = DataManager(
    dataset=dataset,
    labelled_indices=labelled_ixs.tolist(),
    unlabelled_indices=unlabelled_ixs.tolist(),
    validation_indices=val_ixs.tolist(),
    test_indices=test_ixs.tolist(),
)
\end{minted}
\vspace{-1.5em}
\captionof{listing}{Defining a PyRelationAL data manager for the MNIST dataset.}\label{code:data_manager}
\end{code}

Next, we create a \texttt{ModelManager} to wrap a machine learning model. In this case, we employ a simple CNN model, which is detailed in Appendix \ref{app:cnn_model}. The uncertainty estimator in PyRelational is typically set by the \texttt{ModelManager}. In Listing \ref{code:model_wrapper_definition}, we use the \texttt{LightningMCDropoutModelManager}, which applies MCDropout \cite{gal2016dropout} to estimate uncertainty. Since PyRelational includes a pre-defined \texttt{ModelManager}, we simply need to provide the model and configuration details for both the lightning module and \texttt{Trainer}.

\begin{code}

\begin{minted}[linenos,fontsize=\scriptsize,xleftmargin=0.5cm,numbersep=3pt,frame=lines]{python}

from pyrelational.model_managers \
    import LightningMCDropoutModelManager

trainer_config = {
    "patience": 3,
    "epochs": 100,
    "accelerator": "gpu",
    "use_early_stopping": True,
}
model_config = {"num_classes": 10}
model_manager = LightningMCDropoutModelManager(
    model_class=ConvNet,
    model_config=model_config,
    trainer_config=trainer_config,
)
\end{minted}
\vspace{-1.5em}
\captionof{listing}{Definition of a model manager wrapping around the machine learning model.}\label{code:model_wrapper_definition}

\end{code}

With the model and dataset prepared, we proceed to define our active learning strategy. In Listing \ref{code:strategy}, we use PyRelational's two-step framework to combine the \texttt{LeastConfidence} informativeness score with the \texttt{ProbabilisticSampler}, creating a composite strategy.

\begin{code}

\begin{minted}[linenos,fontsize=\scriptsize,xleftmargin=0.5cm,numbersep=3pt,frame=lines]{python}
from pyrelational.strategies.classification \
    import ClassificationStrategy
from pyrelational.batch_mode_samplers import \
    ProbabilisticSampler
from pyrelational.informativeness import \
    LeastConfidence

strategy = ClassificationStrategy(
    scorer=LeastConfidence(),
    sampler=ProbabilisticSampler(),
)
\end{minted}
\vspace{-1.5em}
\captionof{listing}{Definition of a simple least confidence strategy with probabilistic sampling.}\label{code:strategy}
\end{code}

At this point, most of the components needed for the active learning experiment are in place. The last step is to configure the oracle. For this benchmarking exercise, we utilize PyRelational's \texttt{BenchmarkOracle}, which retrieves labels directly from the dataset. Lastly, we use PyRelational's \texttt{Pipeline} to tie all the components together, as shown in Listing \ref{code:run_example}.  

\begin{code}

\begin{minted}[linenos,fontsize=\scriptsize,xleftmargin=0.5cm,numbersep=3pt,frame=lines]{python}

from pyrelational.oracles import BenchmarkOracle
from pyrelational.pipeline import Pipeline

oracle = BenchmarkOracle()
pipeline = Pipeline(
    data_manager=data_manager,
    model_manager=model_manager,
    strategy=strategy,
    oracle=oracle,
)
pipeline.run(
    num_annotate=10, 
    num_iterations=25,
)

\end{minted}
\vspace{-1.5em}
\captionof{listing}{Running the strategy.}\label{code:run_example}
\end{code}

In summary, this case study illustrates how PyRelational simplifies the construction of an active learning pipeline. By integrating a \texttt{DataManager}, \texttt{ModelManager}, and customisable strategies, users can easily set up and run active learning experiments with minimal effort. Notably, the majority of code required to setup the experiment is in the implementation of the model itself, rather than any PyRelationAL components (85 lines for the model versus 71 lines combined for running the AL experiment).

\section{Datasets and Benchmarking}

A core part of ML research and development is the translation of theory to practice typically in the form of evaluating a proposed method empirically against a range of established or challenging benchmark datasets and tasks. The practice of empirically evaluating on established datasets and tasks has become so ubiquitous that all major ML frameworks such as PyTorch, TensorFlow, SciKit-Learn come with interfaces for downloading common benchmarks. The rapid progress of ML research in images, text, and graphs can be attributed, at least partially, to easy access to pre-processed benchmark datasets \citep{imagenet_cvpr09,wang2019superglue,hu2020open,huang2021therapeutics}.

\begin{table*}[ht]
    \caption{Selection of datasets made available in the PyRelationAL library along with information on whether the ML task is a classification or regression (denoted C or R, respectively), whether it is real-world or synthetic dataset, and the raw source of our implementation or the paper from which we constructed the dataset if it was not available publicly. We also give the licence of each dataset curated and extend our permissive Apache 2.0 licence (see section \ref{sec:maintenance}) to synthetic datasets generated with our code as well as the task configurations. Our selection was strongly based on their prior reference in related AL literature. More details on each of the datasets can be found in Appendix \ref{app:datasetsummaries}. \vspace{0.1cm}}
    \label{tab:aldatasets}
    \begin{center}
    \resizebox{.9\textwidth}{!}{%

\begin{tabular}{@{}llllr@{}}
\toprule  
Name & Type & Example use in AL literature & Licence & Size \\ \midrule
\multicolumn{5}{l}{\centering \textbf{Classification Datasets}} \\ \midrule
BreastCancer \cite{BreastCancerDataset} & Real & \citet{bald} & CC BY 4.0 & 569 \\
Checkerboard2x2 \cite{kseniaLAL} & Synth. & \citet{kseniaLAL} & Apache 2.0 & 2,000 \\
Checkerboard4x4 \cite{kseniaLAL} & Synth. & \citet{kseniaLAL} & Apache 2.0 & 2,000 \\
CreditCard \cite{creditcard} & Real & \citet{kseniaLAL} & DbCL 1.0 & 284,807 \\
FashionMNIST \cite{xiao2017/online} & Real & \citet{kirsch2019batchbald} & MIT & 70,000 \\
GaussianClouds \cite{kseniaLAL} & Synth. & \citet{kseniaLAL} & Apache 2.0 & 11,000 \\
Glass \cite{UCI} & Real & \citet{caiAL} & CC BY 4.0 & 213 \\
MNIST \cite{LeCunMNIST} & Real & \citet{kirsch2019batchbald} & CC BY SA 3.0 & 70,000 \\
Parkinsons \cite{ParkinsonsDataset} & Real & \citet{xiongAL} & CC BY 4.0 & 195 \\
StriatumMini \cite{ogstriatumsource} & Real & \citet{kseniaLAL, kseniaGIStriatum} & GPLv3 & 20,000 \\
SynthClass1 \cite{kseniaLAL} & Synth. & \citet{kseniaLAL} & Apache 2.0 & 500 \\
SynthClass2 \cite{kseniaLAL} & Synth. & \citet{kseniaLAL} & Apache 2.0 & 500 \\ \midrule
\multicolumn{5}{l}{\centering  \textbf{Regression Datasets}} \\ \midrule
Airfoil \cite{UCI} & Real & \citet{dongruiWURegressionAL} & CC BY 4.0 & 1,502 \\
Concrete \cite{concretedataset} & Real & \citet{dongruiWURegressionAL} & CC BY 4.0 & 1,030 \\
Diabetes \cite{diabetesEfron} & Real & \citet{bald} & CC BY 4.0 & 442 \\
DrugComb \cite{drugcombdb} & Real & \citet{bertin2023recover} & CC BY-NC 4.0 & 739,964 \\
Energy \cite{energydataset} & Real & \citet{Pinsler2019BayesianBA} & CC BY 4.0 & 768 \\
Power \cite{powerdataset} & Real & \citet{Pinsler2019BayesianBA} & CC BY 4.0 & 9,568 \\
SynthReg1 (Proposed) & Synth. & This paper & Apache 2.0 & 1,000 \\
SynthReg2 (Proposed) & Synth. & This paper & Apache 2.0 & 1,000 \\
WineQuality \cite{winequalitydataset} & Real & \citet{dongruiWURegressionAL} & CC BY 4.0 & 1,598 \\
Yacht \cite{UCI} & Real & \citet{Pinsler2019BayesianBA} & CC BY 4.0 & 306 \\
\bottomrule
\end{tabular}
}
\end{center}
\end{table*}

However, despite a multitude of papers, surveys, and other libraries for AL methods there is no established set of datasets evaluating AL strategies. To further complicate matters, the datasets can be processed to pose different AL tasks such as cold and warm starts as described in \citet{kseniaLAL} and \citet{yangloog} on top of the usual train, validation, test splits provided to the ML task model. In other words, a benchmark for AL has to be considered from the characteristics of the dataset and the circumstances of the initial labelling subset of the train set. 


Due to these challenges, AL papers have a tendency to test and apply proposed methods using different splits of the same datasets with little overlap agreement across papers --- as noted in several reviews \citep{zhanreview, yangloog, dongruiWURegressionAL}. This makes it difficult to assess strategies across a range of common datasets and identify regimes under which a given strategy shows success or failure. This issue is exacerbated by publication bias which favours the reporting of positive results --- as noted in the seminal review by Settles \citep{pmlr-v16-settles11a}. 

\subsection{Datasets} \label{sec:datasets}

To help alleviate this issue we have collected a variety of datasets for both classification and regression tasks as used in AL literature and make these available through PyRelationAL, see Table \ref{tab:aldatasets}. The datasets are either real world datasets or synthetic datasets taken from AL literature  \citep{kseniaLAL, creditcard, zhanreview, yangloog, dongruiWURegressionAL, Pinsler2019BayesianBA}. The synthetic datasets were typically devised to pose challenges to specific strategies. Another selection criterion was their permissive licensing, such as the Creative Commons Attribution 4.0 International Licence granted on the recently updated UCI archives \citep{UCI}, or the licences through the Open Data Commons initiative.


\subsection{Benchmarking Utilities} \label{sec: benchmarkingsection}

Active learning strategies are evaluated differently from the traditional machine learning models due to the iterative nature and focus on maximising data utility with minimal labelling. Unlike the static task model that is evaluated post-training on the hold-out test set, the AL strategy is assessed over multiple acquisition steps as they query data to improve the task model's performance. Therefore, additional considerations have to be made in the construction of the experiments, reproducibility and reporting to assess the qualities of AL strategies. 

At minimum, one has to consider the initial labelled set provided to the task model. For example, whether the model is given a cold start (initial model trained on minimal or no labelled data) versus a warm start (where the initial model is trained on a more substantially labelled set) \cite{kseniaLAL}. Other contextual factors exist such as assessing whether strategies can find and query out of distribution data, and construct training datasets that are representative of the broader data distribution \cite{monarch_human---loop_2021}. 

Our goals with regard to benchmarking were to provide:

\begin{enumerate}
    \item Datasets that have already been used in AL literature before (see Section \ref{sec:datasets}).
    \item Experiments with publicly available results in raw format online, which can be used to produce figures and results tables directly.
    \item Scripts and utilities that produce the results for reproducibility, and allow users to incorporate their own datasets, AL task initialisations, and AL strategies, in a consistent and approachable format.
\end{enumerate}

In our benchmarking directory, each dataset and corresponding AL benchmark is associated with: a \texttt{DataManager}, a task \texttt{ModelManager}, and an experiment module.

As mentioned previously, the \texttt{DataManager} defines the initial structure of the data subsets. Typically, each dataset is divided into a train, validation, and hold-out test set; when a canonical hold-out test set is available, it is used. Otherwise, random splits are applied for both classification and regression tasks. The DataManager also specifies the initial division of the training set into labelled and unlabelled subsets, from which the active learning strategy queries annotations. When available, we use settings reported directly in the literature; otherwise, we follow warm/cold start setups as described in \citet{kseniaLAL}, with 10\% of the data labelled for warm starts and one annotated observation per class (or one for regression) for cold starts.

\begin{figure*}
\begin{center}
    
    \includegraphics[width=0.9\textwidth]{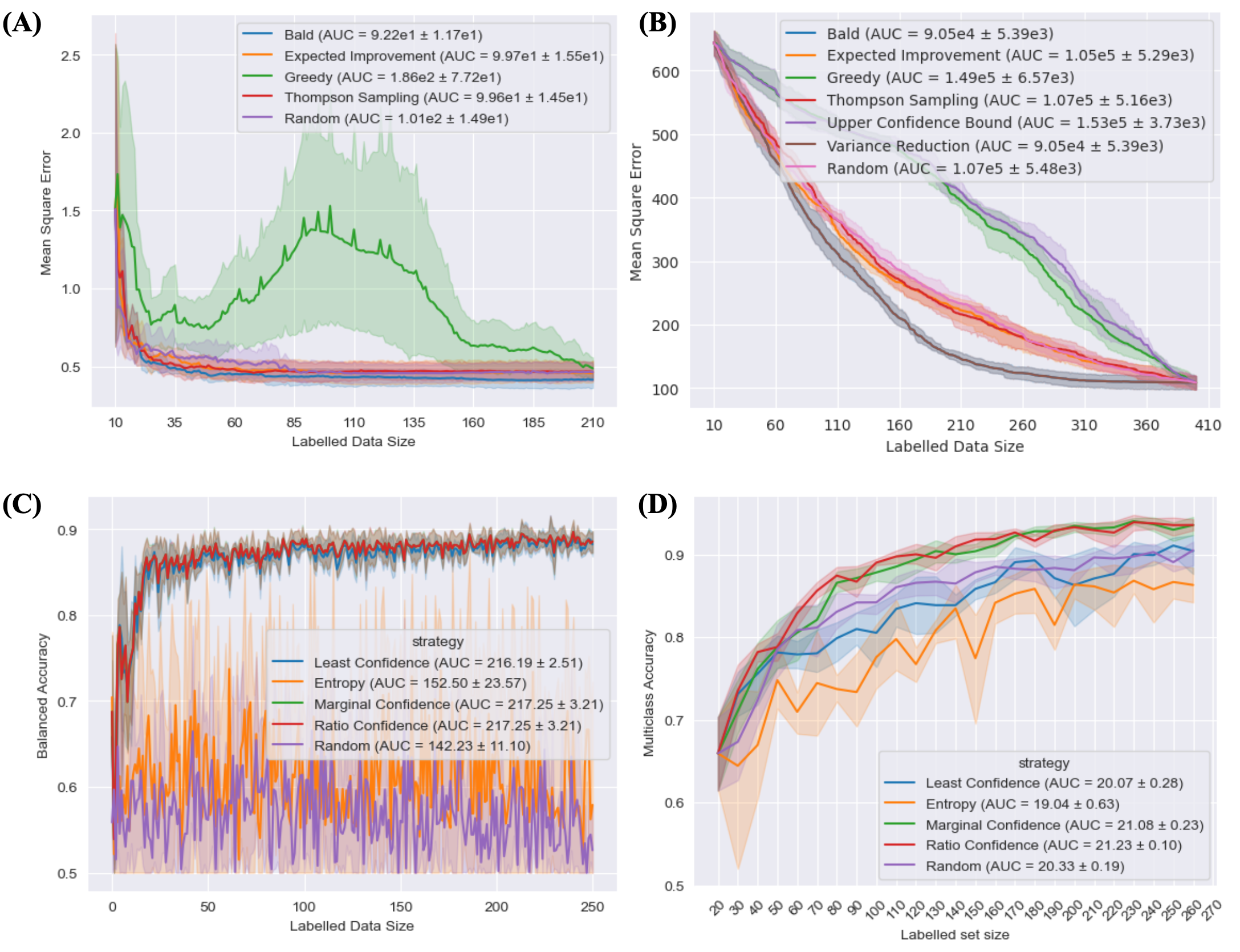}
\end{center}
    \caption{Four samples of benchmark results obtained for regression (top row) and classification (bottom row) scenarios. Note the different initialisations, single or batch-mode acquisition, and task models, being utilised for the strategies. (A) SynthReg2 dataset, regression with MLP ensemble with bagging, warm start, single acquisition. (B) Energy, regression with GP, cold start, single acquisition. (C) CreditCard, binary classification with random forest, warm start, single acquisition. (D) MNIST, multi-class classification with CNN+MCDropOut, warm start, Top-K batch acquisition of K=10. In each panel, the legend reports the area under the curve as a global metric. It is important to remark that depending on the metric, a low AUC might be preferable, in particular for regression where we measure by mean square error we favour a lower AUC for the metric.}
    \label{fig:benchmarksample}
\end{figure*}

The task model manager specifies the class of model used in the task. We use established models for tasks typically associated with a model or previously used in AL literature, such as CNNs with MCDropout for MNIST \cite{kirsch2019batchbald}. For our experiments in Figure \ref{fig:benchmarksample}, we have used the class balanced accuracy for classification tasks and mean square error in regression on the hold-out test set. However, this is not necessarily applied to all benchmarks and the task model along with relevant metrics for evaluation should be adapted to the context of the task at hand.

Finally, the experiment script is composed of a trial definition and an experiment manager. The trial defines the actions to be performed upon a single configuration of the strategy to be applied in the initial settings described by the DataManager and ModelManager, controlled by set seeds. Each configuration is replicated multiple times under set seeds for reproducibility, and all of the trials along with the hardware allocated to run the trials is managed by the experiment manager. This allows the user to rerun experiments according to the hardware available to them and make best use of parallelisation. The outputs of the experiments and experiment parameters are collected into a simple CSV file which can be used to analyse and plot the results as shown in Figure \ref{fig:benchmarksample}.

Figure \ref{fig:benchmarksample} shows a sample of strategies applied to regression and classification scenarios (top and bottom) under different evaluation setups. Overall, the results highlight the economic value of applying active learning --- as in all scenarios at least one non-random strategy performs significantly better than random acquisition. In particular, in Figure 2C we see that the Random and Entropy strategies do not improve the MLP classifier model in the credit card dataset under the alloted number of annotations. Similarly in Figure 2B, we see the utility of the variance reduction strategy over the other strategies. Despite the similarities between the uncertainty based sampling algorithms least confidence, marginal confidence, ratio confidence (see Appendix \ref{appendix:infos}) and in Figure 2C, we see that these can bring about significant differences in Figure 2D, particularly when we consider batch selections. These differences highlight the value of exploring a variety active learning strategies across a variety of different active learning task scenarios. 

While our current benchmarks focus on the impact of different strategies, keeping everything else fixed, we believe that a successful active learning solution would consider all other parameters including the initial seed, the model, and the uncertainty estimator (if relevant). As such, the only thing that should remain fixed in benchmarking is the held-out test set. We will continue to add and expand benchmarks accounting for these consideration, and we provide here utilities for the community to engage in this effort. Particularly, we will put a priority on moving away from smaller classically well-balanced UCI datasets often found in older AL literature, to larger high-dimensional datasets with imbalanced classes and other modalities such as graphs, as with DrugComb \cite{drugcombdb}.

\section{Discussion and comparison on existing open-source libraries}

\begin{table*}
    \caption{Open source libraries for AL and features available alongside their open source licence used. \textsuperscript{(*)} scikit-activeml and ModAL can work with models defined in Keras and PyTorch via wrapping dependency to Skorch \cite{skorch} which provides them with an interface akin to Scikit-learn estimators.\textsuperscript{($\dagger$)} AliPy lists a planned update for regression strategies. For uncertainty estimation, we looked at whether the toolkit has modules for estimating uncertainty in point estimate models either through ensembling or methods such as MCDropout or SWAG.}
    \label{tab:libraries-summary}
    \resizebox{\textwidth}{!}{%
    \begin{tabular}{@{}llllllll@{}}
    \toprule
    Name (Year) & \begin{tabular}[c]{@{}l@{}}Main ML \\ framework\end{tabular} & \begin{tabular}[c]{@{}l@{}}Strategies\\ Classification\end{tabular} & \begin{tabular}[c]{@{}l@{}}Strategies\\ Regression\end{tabular} & \begin{tabular}[c]{@{}l@{}} Uncertainty \\ estimation \end{tabular} & Datasets & \begin{tabular}[c]{@{}l@{}} Benchmarking \\ utilities \end{tabular} & Licence \\ \midrule
    \begin{tabular}[c]{@{}l@{}}JCLAL \cite{jclal} \end{tabular} & \begin{tabular}[c]{@{}l@{}}Weka/MULAN\end{tabular}  & \cmark & \xmark & \xmark & \xmark & \xmark & GPL 3.0 \\
    \begin{tabular}[c]{@{}l@{}}AliPy \cite{alipy}\end{tabular} & Scikit-learn  & \cmark & \xmark\textsuperscript{($\dagger$)} & \xmark & \xmark & \xmark & BSD-3-Clause \\
    \begin{tabular}[c]{@{}l@{}}LibAcT \cite{libact}\end{tabular} & Scikit-learn & \cmark & \xmark & \xmark & \xmark & \xmark & BSD-2-Clause \\
    \begin{tabular}[c]{@{}l@{}}ModAL \cite{modAL2018}\end{tabular} & Scikit-learn\textsuperscript{(*)} &  \cmark & \cmark & \cmark & \xmark & \xmark & MIT \\
    \begin{tabular}[c]{@{}l@{}}BAAL \cite{baal}\end{tabular} & PyTorch &  \cmark & \cmark & \cmark & \xmark & \xmark & Apache 2.0 \\
    \begin{tabular}[c]{@{}l@{}}scikit-activeml \cite{skactiveml2021}\end{tabular} & Scikit-learn\textsuperscript{(*)} & \cmark & \cmark & \cmark & \xmark & \xmark & BSD-3-Clause \\
    \begin{tabular}[c]{@{}l@{}}CardinAL \cite{cardinal}\end{tabular} & Scikit-learn &  \cmark & \xmark & \xmark & \xmark & \xmark & Apache 2.0 \\
    \midrule
    \begin{tabular}[c]{@{}l@{}}\textbf{PyRelationAL (Proposed)} \end{tabular} & \begin{tabular}[c]{@{}l@{}}\textbf{Generic/PyTorch}\end{tabular} & \cmark & \cmark & \textbf{\cmark} & \cmark & \cmark & \textbf{Apache 2.0} \\ \bottomrule
    \end{tabular}%
    }
\end{table*}

One of the main hurdles in active learning research on strategies is the software engineering overhead associated with constructing robust AL infrastructure \citep{baldridge-palmer-2009-well, pmlr-v16-settles11a}. AL systems have to interface robustly between data managers, annotators, machine learning estimators. This is no simple task, as we also want them to be extendable and easy to use. Several open source software packages have been made to address this problem, but tackle the issues across different fronts as shown in Table \ref{tab:libraries-summary}.

PyRelationAL differentiates itself across 4 main points which we find additive to the AL software landscape:

\begin{enumerate}
    \item \textbf{Strategy design framework}: as described in Section \ref{ssec:strat_design}, a key part of our library is the construction of single- and batch-acquisition strategies via a two step framework that allows users to quickly construct existing and novel strategies. This is not an aspect found in the other toolkits.
    \item \textbf{ML framework agnostic infrastructure}: Unlike most libraries, the proposed framework is agnostic to the ML framework being used. This allows for greater flexibility, not only to the choice of model performing the ML task of interest, but also in approach to modelling uncertainty especially important to uncertainty based active learning strategies. We recognise this comes with additional overhead, but judge that this is minimal compared to the flexibility provided. 
    \item \textbf{Datasets}: PyRelationAL provides an interface for the download and wrapping of commonly used (and appropriately licensed) datasets for quick experimentation and benchmarking for both classification and regression tasks. Tutorials are available describing the construction of DataManagers for bespoke datasets as well.
    \item \textbf{Benchmarking utilities}: In addition to our core modules, we have prepared benchmarking utilities to compare existing and novel strategies on library and bespoke datasets in a controlled and consistent manner, to address one of the main empirical challenges facing this field. This is also not available in any of the other packages. 
\end{enumerate}

\subsection{Strategy design framework}

Strategy implementations vary significantly across libraries, often influenced by the ML framework used for the task models. Some libraries, like JCLAL, AliPy, LibAcT, and CardinAL, focus specifically on classification-based AL strategies, which have seen more direct research in active learning \citep{burrsettlesbook}. In contrast, regression scenarios are more commonly addressed in fields such as experimental design and statistics \citep{kiefer1959, fedorov1972theory}. scikit-activeml \cite{skactiveml2021} offers the broadest coverage, with 42 strategies spanning both classification and regression and supporting pool-based and streaming scenarios. 

The diversity in strategy offerings is not only driven by the ML task type but also by the active learning paradigm, such as pool-based, streaming-based, or generative-based sampling. For instance, scikit-activeml and BAAL support both pool and streaming-based AL, while other libraries, including PyRelationAL, focus solely on pool-based sampling. Additional problem constraints, such as noisy oracles and cost-sensitive labelling, further diversify the strategy implementations across libraries. AliPy, for example, includes strategies specifically designed for these scenarios \cite{alipy}. 

Currently, PyRelationAL supports 15 informativeness measures and 2 batch sampling algorithms across classification and regression, yielding about 30 strategies (some combinations are redundant) that can be paired with 4 uncertainty estimation methods, including MCDropout for neural networks, model committees, bagging, and bootstrap ensembles. We plan to expand this, particularly in the area of batch selection algorithms. Nonetheless, the current version covers most foundational strategies described by Settles \cite{burrsettlesbook}, Federov \cite{fedorov1972theory}, and Monarch \cite{monarch_human---loop_2021}, as well as seminal approaches such as BALD \cite{bald}.

As discussed, a key advantage of PyRelationAL in AL strategy definition is its modular two-step framework, which embodies strong software design by clearly separating responsibilities for scoring and sampling. This compositional approach allows for flexible and interchangeable strategy components. Among comparable libraries, only BAAL offers a similar functionality through its \texttt{StochasticHeuristic} class, which wraps around deterministic heuristics to alter sampling strategies. However, BAAL does not explicitly separate scoring from sampling, limiting its flexibility in comparison.

\subsection{ML framework agnostic infrastructure}

Most libraries are designed around a particular ML framework, with strategies that align closely with the model types available in that framework. For instance, scikit-activeml \citep{skactiveml2021} and ModAL \cite{modAL2018} primarily use scikit-learn estimators, with both libraries supporting deep learning packages like PyTorch through a Skorch dependency \citep{skorch}. BAAL \citep{baal} is focused on Bayesian uncertainty-based active learning strategies tailored for deep learning models in PyTorch, leveraging wrappers such as MCDropout \cite{gal2016dropout} to estimate model uncertainties. AliPy \citep{alipy}, on the other hand, centers its strategies around scikit-learn’s logistic regressor, although it can interface with other frameworks if model implementations adhere to the scikit-learn API, as noted in their GitHub issues \cite{miscissue2alipy}.

In contrast, PyRelationAL is designed with ML framework-agnostic interfaces, offering flexibility across model types and frameworks. This approach enables active learning strategies for uncertainty estimation on point-estimate models, such as ensembling, without confining users to a specific model type or ML framework. While the core abstractions are framework-neutral, we include utilities like MCDropout specifically for PyTorch-based models within the library.

Though being ML framework-agnostic may introduce some implementation overhead for users, we view this level of control as essential for research. It allows for the development of strategies that leverage model parameters and supports adaptation to emerging methods—such as epistemic neural networks \citep{epistemicneuralnetworks}—that may not yet have official implementations in major ML frameworks.

\subsection{Datasets and benchmarking}

A critical differentiator and contribution of PyRelationAL is its inclusion of a curated set of datasets, accessible directly through the toolkit, and benchmarking utilities present on the repository and tailored specifically for active learning (AL). These resources facilitate the construction of a publicly available benchmark collection that covers methods implemented in our library for pool-based classification and regression, supporting both single- and batch-mode acquisition. As detailed in Section \ref{sec: benchmarkingsection}, the included scripts and utilities allow users to implement and benchmark their strategies fairly under controlled, reproducible conditions. Users can also leverage existing public results, minimizing the need to re-run computationally intensive AL cycles, especially in deep active learning contexts. These resources directly address key challenges in achieving consistent empirical evaluation across methods, as emphasized in prior literature \cite{pmlr-v16-settles11a, yangloog, zhanreview}.

As shown in Table \ref{tab:libraries-summary}, PyRelationAL stands out as the only package to integrate both of these essential components, supporting AL research and development progress.

\section{Maintaining PyRelationAL} \label{sec:maintenance}

We adopted modern open source software engineering practices in order to promote robustness and reliability of the package whilst maintaining an open policy towards contributions over the long term vision we have for the project. 

\paragraph{Open sourcing, package indexing, and contributing}

PyRelationAL is made available open source under a permissive Apache 2.0 licence with stable releases made on the Python Package Index (PyPI) for easy installation. To introduce new users to the package we created an extensive set of examples, tutorials, and supplementary materials linked through the main README including developer setups for contributors. We follow an inclusive code of conduct for contributors and users to facilitate fair treatment and guidelines for contributing. Furthermore, a contributors guide is provided describing the expected workflows, testing, and code quality expectations for pull requests into the package.

\paragraph{Documentation}

All modules are accompanied by restructured text (ResT) docstrings and comments to maintain documentation of the library. This approach maintains an up to date reference of the API, which is compiled and rendered via Sphinx to be hosted on Read The Docs for every update accepted into the main branch. In addition to the API reference, our documentation website is accompanied by various supplementary tutorials, examples, and information about AL within the library to help get users started and advanced users to extend the library to fit their needs and research endeavours.

\paragraph{Code quality, testing and continuous integration}

To ensure consistent code quality throughout the package, we ensure that source code is PEP-8 compliant via a linter that is checked automatically before a pull request can be approved. We make pre-commit hooks available that can format the code using tools, such as the Black package. All important modules and classes are extensively tested through unit tests to ensure consistent behaviour across versions and operating systems. Code coverage reports are available on the repository as a measure of the health of the package. A continuous integration setup facilitated through Github Actions ensures that any update to the code base is tested across different environments and can be installed reliably.

\section{Conclusion}

PyRelationAL supports popular modelling frameworks and flexibly accounts for the various components within an AL pipeline. Our two step framework allows characterisation and definition of existing and novel active learning strategies that can be deployed against the datasets and active learning tasks which make part of our contributions and are not found in comparable software offerings. We provide detailed documentation and various tutorials across use cases. Furthermore, we adopt modern software practices with an inclusive code of conduct as discussed in Section \ref{sec:maintenance} to foster a productive, sustainable and healthy growth of the library and the community around it. We endeavoured to release PyRelationAL as both a tool to help practitioners perform research but also help newcomers join the community and make novel contributions to the field. We therefore believe PyRelationAL offers a compelling set of features that are additive and beneficial to the AL community.

Our contributions are best seen as a live resource which we will update and expand upon over time. A public list of planned developments is available on the repository, with initial future developments focusing on the development of more batch selection algorithms to rapidly expand our list of strategies. In the longer term, we intend to use the benchmarking utilities we have constructed and further explore the nuances in the data initialisation, batching, model choices, and strategy implementations that challenge active learning strategies and foster innovation in this research area.

\nocite{langley00}

\bibliography{references}

\begin{thebibliography}{}

\bibitem[Abdelwahab and Busso(2019)Abdelwahab and Busso]{abdelwahab2019active}
Abdelwahab, M. and Busso, C. (2019).
\newblock Active learning for speech emotion recognition using deep neural network.
\newblock In {\em 2019 8th International Conference on Affective Computing and Intelligent Interaction (ACII)\/}, pages 1--7. IEEE.

\bibitem[Abraham and Dreyfus-Schmidt(2022)Abraham and Dreyfus-Schmidt]{cardinal}
Abraham, A. and Dreyfus-Schmidt, L. (2022).
\newblock Cardinal, a metric-based active learning framework.
\newblock {\em Software Impacts\/}, {\bf 12}, 100250.

\bibitem[Atighehchian {\em et~al.}(2020)Atighehchian, Branchaud{-}Charron, and Lacoste]{baal}
Atighehchian, P., Branchaud{-}Charron, F., and Lacoste, A. (2020).
\newblock Bayesian active learning for production, a systematic study and a reusable library.
\newblock {\em CoRR\/}, {\bf abs/2006.09916}.

\bibitem[Baldridge and Palmer(2009)Baldridge and Palmer]{baldridge-palmer-2009-well}
Baldridge, J. and Palmer, A. (2009).
\newblock How well does active learning \textit{actually} work? {T}ime-based evaluation of cost-reduction strategies for language documentation.
\newblock In P.~Koehn and R.~Mihalcea, editors, {\em Proceedings of the 2009 Conference on Empirical Methods in Natural Language Processing\/}, pages 296--305, Singapore. Association for Computational Linguistics.

\bibitem[Bertin {\em et~al.}(2023)Bertin, Rector-Brooks, Sharma, Gaudelet, Anighoro, Gross, Mart{\'\i}nez-Pe{\~n}a, Tang, Suraj, Regep, {\em et~al.}]{bertin2023recover}
Bertin, P., Rector-Brooks, J., Sharma, D., Gaudelet, T., Anighoro, A., Gross, T., Mart{\'\i}nez-Pe{\~n}a, F., Tang, E.~L., Suraj, M., Regep, C., {\em et~al.} (2023).
\newblock Recover identifies synergistic drug combinations in vitro through sequential model optimization.
\newblock {\em Cell Reports Methods\/}, {\bf 3}(10).

\bibitem[Brinker(2003)Brinker]{brinkerdiversityanduncertainty}
Brinker, K. (2003).
\newblock Incorporating diversity in active learning with support vector machines.
\newblock In {\em Proceedings of the Twentieth International Conference on International Conference on Machine Learning\/}, ICML'03, page 59–66. AAAI Press.

\bibitem[Brochu {\em et~al.}(2010)Brochu, Cora, and De~Freitas]{brochu2010tutorial}
Brochu, E., Cora, V.~M., and De~Freitas, N. (2010).
\newblock A tutorial on bayesian optimization of expensive cost functions, with application to active user modeling and hierarchical reinforcement learning.
\newblock {\em arXiv preprint arXiv:1012.2599\/}.

\bibitem[Cai {\em et~al.}(2013a)Cai, Zhang, and Zhou]{cai2013maximizing}
Cai, W., Zhang, Y., and Zhou, J. (2013a).
\newblock Maximizing expected model change for active learning in regression.
\newblock In {\em 2013 IEEE 13th international conference on data mining\/}, pages 51--60. IEEE.

\bibitem[Cai {\em et~al.}(2013b)Cai, Zhang, and Zhou]{caiAL}
Cai, W., Zhang, Y., and Zhou, J. (2013b).
\newblock Maximizing expected model change for active learning in regression.
\newblock In {\em 2013 IEEE 13th International Conference on Data Mining\/}, pages 51--60.

\bibitem[Cai {\em et~al.}(2015)Cai, Zhang, and Zhang]{cai2015active}
Cai, W., Zhang, M., and Zhang, Y. (2015).
\newblock Active learning for ranking with sample density.
\newblock {\em Information Retrieval Journal\/}, {\bf 18}, 123--144.

\bibitem[Cohn {\em et~al.}(1994)Cohn, Atlas, and Ladner]{Cohn1994}
Cohn, D., Atlas, L., and Ladner, R. (1994).
\newblock Improving generalization with active learning.
\newblock {\em Machine Learning\/}, {\bf 15}(2), 201--221.

\bibitem[Cortez {\em et~al.}(2009)Cortez, Cerdeira, Almeida, Matos, and Reis]{winequalitydataset}
Cortez, P., Cerdeira, A., Almeida, F., Matos, T., and Reis, J. (2009).
\newblock Modeling wine preferences by data mining from physicochemical properties.
\newblock {\em Decision Support Systems\/}, {\bf 47}(4), 547--553.
\newblock Smart Business Networks: Concepts and Empirical Evidence.

\bibitem[Danka and Horvath(2018)Danka and Horvath]{modAL2018}
Danka, T. and Horvath, P. (2018).
\newblock modal: A modular active learning framework for python.
\newblock {\em arXiv preprint arXiv:1805.00979\/}.

\bibitem[De~Ath {\em et~al.}(2021)De~Ath, Everson, Rahat, and Fieldsend]{de2021greed}
De~Ath, G., Everson, R.~M., Rahat, A.~A., and Fieldsend, J.~E. (2021).
\newblock Greed is good: Exploration and exploitation trade-offs in bayesian optimisation.
\newblock {\em ACM Transactions on Evolutionary Learning and Optimization\/}, {\bf 1}(1), 1--22.

\bibitem[Deng {\em et~al.}(2009)Deng, Dong, Socher, Li, Li, and Fei-Fei]{imagenet_cvpr09}
Deng, J., Dong, W., Socher, R., Li, L.-J., Li, K., and Fei-Fei, L. (2009).
\newblock {ImageNet: A Large-Scale Hierarchical Image Database}.
\newblock In {\em Computer Vision and Pattern Recognition\/}.

\bibitem[Der~Kiureghian and Ditlevsen(2009)Der~Kiureghian and Ditlevsen]{der2009aleatory}
Der~Kiureghian, A. and Ditlevsen, O. (2009).
\newblock Aleatory or epistemic? does it matter?
\newblock {\em Structural Safety\/}, {\bf 31}(2), 105--112.

\bibitem[Dua and Graff(2017)Dua and Graff]{UCI}
Dua, D. and Graff, C. (2017).
\newblock {UCI} machine learning repository.

\bibitem[Efron {\em et~al.}(2004)Efron, Hastie, Johnstone, and Tibshirani]{diabetesEfron}
Efron, B., Hastie, T., Johnstone, I., and Tibshirani, R. (2004).
\newblock {Least angle regression}.
\newblock {\em The Annals of Statistics\/}, {\bf 32}(2), 407 -- 499.

\bibitem[Fedorov(1972)Fedorov]{fedorov1972theory}
Fedorov, V. (1972).
\newblock {\em Theory of Optimal Experiments\/}.
\newblock Cellular Neurobiology. Academic Press.

\bibitem[Gal and Ghahramani(2016)Gal and Ghahramani]{gal2016dropout}
Gal, Y. and Ghahramani, Z. (2016).
\newblock Dropout as a bayesian approximation: Representing model uncertainty in deep learning.
\newblock In {\em International Conference on Machine Learning\/}, pages 1050--1059. PMLR.

\bibitem[Gaudelet {\em et~al.}(2021)Gaudelet, Day, Jamasb, Soman, Regep, Liu, Hayter, Vickers, Roberts, Tang, Roblin, Blundell, Bronstein, and Taylor-King]{gaudelet_drugs}
Gaudelet, T., Day, B., Jamasb, A.~R., Soman, J., Regep, C., Liu, G., Hayter, J. B.~R., Vickers, R., Roberts, C., Tang, J., Roblin, D., Blundell, T.~L., Bronstein, M.~M., and Taylor-King, J.~P. (2021).
\newblock Utilizing graph machine learning within drug discovery and development.
\newblock {\em Briefings in Bioinformatics\/}, {\bf 22}(6).
\newblock bbab159.

\bibitem[Github(2019)Github]{miscissue2alipy}
Github (2019).
\newblock Using alipy with tensorflow.
\newblock \url{https://github.com/NUAA-AL/ALiPy/issues/2}.
\newblock Accessed: 2024-09-30.

\bibitem[Guo and Schuurmans(2007)Guo and Schuurmans]{guobatchmode}
Guo, Y. and Schuurmans, D. (2007).
\newblock Discriminative batch mode active learning.
\newblock In {\em Proceedings of the 20th International Conference on Neural Information Processing Systems\/}, NIPS'07, page 593–600, Red Hook, NY, USA. Curran Associates Inc.

\bibitem[Hie {\em et~al.}(2020)Hie, Bryson, and Berger]{hie2020leveraging}
Hie, B., Bryson, B.~D., and Berger, B. (2020).
\newblock Leveraging uncertainty in machine learning accelerates biological discovery and design.
\newblock {\em Cell Systems\/}, {\bf 11}(5), 461--477.

\bibitem[Hoi {\em et~al.}(2006)Hoi, Jin, Zhu, and Lyu]{hoi2006batch}
Hoi, S.~C., Jin, R., Zhu, J., and Lyu, M.~R. (2006).
\newblock Batch mode active learning and its application to medical image classification.
\newblock In {\em Proceedings of the 23rd International Conference on Machine Learning\/}, pages 417--424.

\bibitem[Houlsby {\em et~al.}(2011)Houlsby, Husz{\'a}r, Ghahramani, and Lengyel]{bald}
Houlsby, N., Husz{\'a}r, F., Ghahramani, Z., and Lengyel, M. (2011).
\newblock Bayesian active learning for classification and preference learning.
\newblock {\em arXiv preprint arXiv:1112.5745\/}.

\bibitem[Hu {\em et~al.}(2020)Hu, Fey, Zitnik, Dong, Ren, Liu, Catasta, and Leskovec]{hu2020open}
Hu, W., Fey, M., Zitnik, M., Dong, Y., Ren, H., Liu, B., Catasta, M., and Leskovec, J. (2020).
\newblock Open graph benchmark: Datasets for machine learning on graphs.
\newblock {\em Advances in Neural Information Processing Systems\/}, {\bf 33}, 22118--22133.

\bibitem[Huang {\em et~al.}(2021)Huang, Fu, Gao, Zhao, Roohani, Leskovec, Coley, Xiao, Sun, and Zitnik]{huang2021therapeutics}
Huang, K., Fu, T., Gao, W., Zhao, Y., Roohani, Y., Leskovec, J., Coley, C.~W., Xiao, C., Sun, J., and Zitnik, M. (2021).
\newblock Therapeutics data commons: Machine learning datasets and tasks for drug discovery and development.
\newblock {\em arXiv preprint arXiv:2102.09548\/}.

\bibitem[Kiefer(1959)Kiefer]{kiefer1959}
Kiefer, J. (1959).
\newblock Optimum experimental designs.
\newblock {\em Journal of the Royal Statistical Society. Series B (Methodological)\/}, {\bf 21}(2), 272--319.

\bibitem[Kirsch {\em et~al.}(2019)Kirsch, Van~Amersfoort, and Gal]{kirsch2019batchbald}
Kirsch, A., Van~Amersfoort, J., and Gal, Y. (2019).
\newblock Batchbald: Efficient and diverse batch acquisition for deep bayesian active learning.
\newblock {\em Advances in Neural Information Processing Systems\/}, {\bf 32}.

\bibitem[Kirsch {\em et~al.}(2023)Kirsch, Farquhar, Atighehchian, Jesson, Branchaud-Charron, and Gal]{kirsch2023stochastic}
Kirsch, A., Farquhar, S., Atighehchian, P., Jesson, A., Branchaud-Charron, F., and Gal, Y. (2023).
\newblock Stochastic batch acquisition: A simple baseline for deep active learning.
\newblock {\em Transactions on Machine Learning Research\/}.
\newblock Expert Certification.

\bibitem[Konyushkova {\em et~al.}(2015)Konyushkova, Sznitman, and Fua]{kseniaGIStriatum}
Konyushkova, K., Sznitman, R., and Fua, P.~V. (2015).
\newblock Introducing geometry in active learning for image segmentation.
\newblock {\em 2015 IEEE International Conference on Computer Vision (ICCV)\/}, pages 2974--2982.

\bibitem[Konyushkova {\em et~al.}(2017)Konyushkova, Sznitman, and Fua]{kseniaLAL}
Konyushkova, K., Sznitman, R., and Fua, P. (2017).
\newblock Learning active learning from data.
\newblock {\em Advances in Neural Information Processing Systems\/}, {\bf 30}.

\bibitem[Kottke {\em et~al.}(2021)Kottke, Herde, Minh, Benz, Mergard, Roghman, Sandrock, and Sick]{skactiveml2021}
Kottke, D., Herde, M., Minh, T.~P., Benz, A., Mergard, P., Roghman, A., Sandrock, C., and Sick, B. (2021).
\newblock scikitactiveml: A library and toolbox for active learning algorithms.
\newblock {\em Preprints\/}.

\bibitem[Lakshminarayanan {\em et~al.}(2017)Lakshminarayanan, Pritzel, and Blundell]{lakshminarayanan_simple_2017}
Lakshminarayanan, B., Pritzel, A., and Blundell, C. (2017).
\newblock Simple and {Scalable} {Predictive} {Uncertainty} {Estimation} using {Deep} {Ensembles}.
\newblock {\em arXiv:1612.01474 [cs, stat]\/}.
\newblock arXiv: 1612.01474.

\bibitem[Landrum {\em et~al.}(2024)Landrum, Tosco, Kelley, Rodriguez, Cosgrove, Vianello, sriniker, Gedeck, Jones, NadineSchneider, Kawashima, Nealschneider, Dalke, Swain, Cole, Turk, Savelev, Vaucher, Wójcikowski, Take, Scalfani, Walker, Probst, Ujihara, tadhurst cdd, Pahl, guillaume godin, Lehtivarjo, Bérenger, and Bisson]{rdkit}
Landrum, G., Tosco, P., Kelley, B., Rodriguez, R., Cosgrove, D., Vianello, R., sriniker, Gedeck, P., Jones, G., NadineSchneider, Kawashima, E., Nealschneider, D., Dalke, A., Swain, M., Cole, B., Turk, S., Savelev, A., Vaucher, A., Wójcikowski, M., Take, I., Scalfani, V.~F., Walker, R., Probst, D., Ujihara, K., tadhurst cdd, Pahl, A., guillaume godin, Lehtivarjo, J., Bérenger, F., and Bisson, J. (2024).
\newblock rdkit/rdkit: 2024\_09\_2 (q3 2024) release.

\bibitem[Lecun {\em et~al.}(1998)Lecun, Bottou, Bengio, and Haffner]{LeCunMNIST}
Lecun, Y., Bottou, L., Bengio, Y., and Haffner, P. (1998).
\newblock Gradient-based learning applied to document recognition.
\newblock {\em Proceedings of the IEEE\/}, {\bf 86}(11), 2278--2324.

\bibitem[Little {\em et~al.}(2007)Little, McSharry, Roberts, Costello, and Moroz]{ParkinsonsDataset}
Little, M.~A., McSharry, P.~E., Roberts, S.~J., Costello, D.~A., and Moroz, I.~M. (2007).
\newblock Exploiting nonlinear recurrence and fractal scaling properties for voice disorder detection.
\newblock {\em BioMedical Engineering OnLine\/}, {\bf 6}(1), 23.

\bibitem[Lucchi {\em et~al.}(2013)Lucchi, Li, and Fua]{ogstriatumsource}
Lucchi, A., Li, Y., and Fua, P. (2013).
\newblock Learning for structured prediction using approximate subgradient descent with working sets.
\newblock In {\em 2013 IEEE Conference on Computer Vision and Pattern Recognition\/}, pages 1987--1994.

\bibitem[Mehrjou {\em et~al.}(2021)Mehrjou, Soleymani, Jesson, Notin, Gal, Bauer, and Schwab]{mehrjou_genedisco_2021}
Mehrjou, A., Soleymani, A., Jesson, A., Notin, P., Gal, Y., Bauer, S., and Schwab, P. (2021).
\newblock {GeneDisco}: {A} {Benchmark} for {Experimental} {Design} in {Drug} {Discovery}.
\newblock {\em arXiv:2110.11875 [cs, stat]\/}.
\newblock arXiv: 2110.11875.

\bibitem[Monarch(2021)Monarch]{monarch_human---loop_2021}
Monarch, R.~M. (2021).
\newblock {\em Human-{In}-the-{Loop} {Machine} {Learning}: {Active} {Learning} and {Annotation} for {Human}-{Centered} {AI}.}
\newblock Manning Publications Co. LLC.

\bibitem[Osband {\em et~al.}(2024)Osband, Wen, Asghari, Dwaracherla, Ibrahimi, Lu, and Van~Roy]{epistemicneuralnetworks}
Osband, I., Wen, Z., Asghari, S.~M., Dwaracherla, V., Ibrahimi, M., Lu, X., and Van~Roy, B. (2024).
\newblock Epistemic neural networks.
\newblock In {\em Proceedings of the 37th International Conference on Neural Information Processing Systems\/}, NIPS '23, Red Hook, NY, USA. Curran Associates Inc.

\bibitem[Pinsler {\em et~al.}(2019)Pinsler, Gordon, Nalisnick, and Hern{\'a}ndez-Lobato]{Pinsler2019BayesianBA}
Pinsler, R., Gordon, J., Nalisnick, E., and Hern{\'a}ndez-Lobato, J.~M. (2019).
\newblock Bayesian batch active learning as sparse subset approximation.
\newblock {\em Advances in Neural Information Processing Systems\/}, {\bf 32}.

\bibitem[Pozzolo {\em et~al.}(2015)Pozzolo, Caelen, Johnson, and Bontempi]{creditcard}
Pozzolo, A.~D., Caelen, O., Johnson, R.~A., and Bontempi, G. (2015).
\newblock Calibrating probability with undersampling for unbalanced classification.
\newblock In {\em 2015 IEEE Symposium Series on Computational Intelligence\/}, pages 159--166.

\bibitem[Ren {\em et~al.}(2021)Ren, Xiao, Chang, Huang, Li, Gupta, Chen, and Wang]{ren2021survey}
Ren, P., Xiao, Y., Chang, X., Huang, P.-Y., Li, Z., Gupta, B.~B., Chen, X., and Wang, X. (2021).
\newblock A survey of deep active learning.
\newblock {\em ACM Computing Surveys (CSUR)\/}, {\bf 54}(9), 1--40.

\bibitem[Reyes {\em et~al.}(2016)Reyes, P{{\'e}}rez, del Carmen Rodr{{\'i}}guez-Hern{{\'a}}ndez, Fardoun, and Ventura]{jclal}
Reyes, O., P{{\'e}}rez, E., del Carmen Rodr{{\'i}}guez-Hern{{\'a}}ndez, M., Fardoun, H.~M., and Ventura, S. (2016).
\newblock Jclal: A java framework for active learning.
\newblock {\em Journal of Machine Learning Research\/}, {\bf 17}(95), 1--5.

\bibitem[Russo {\em et~al.}(2018)Russo, Van~Roy, Kazerouni, Osband, Wen, {\em et~al.}]{russo2018tutorial}
Russo, D.~J., Van~Roy, B., Kazerouni, A., Osband, I., Wen, Z., {\em et~al.} (2018).
\newblock A tutorial on thompson sampling.
\newblock {\em Foundations and Trends in Machine Learning\/}, {\bf 11}(1), 1--96.

\bibitem[Settles(2011)Settles]{pmlr-v16-settles11a}
Settles, B. (2011).
\newblock From theories to queries: Active learning in practice.
\newblock In I.~Guyon, G.~Cawley, G.~Dror, V.~Lemaire, and A.~Statnikov, editors, {\em Active Learning and Experimental Design workshop In conjunction with AISTATS 2010\/}, volume~16 of {\em Proceedings of Machine Learning Research\/}, pages 1--18, Sardinia, Italy. PMLR.

\bibitem[Settles(2012)Settles]{burrsettlesbook}
Settles, B. (2012).
\newblock Uncertainty sampling.
\newblock In {\em Active Learning\/}, Synthesis Lectures on Artificial Intelligence and Machine Learning, pages 11--21. Morgan \& Claypool Publishers.

\bibitem[Seung {\em et~al.}(1992a)Seung, Opper, and Sompolinsky]{seung1992query}
Seung, H.~S., Opper, M., and Sompolinsky, H. (1992a).
\newblock Query by committee.
\newblock In {\em Proceedings of the fifth annual workshop on Computational learning theory\/}, pages 287--294.

\bibitem[Seung {\em et~al.}(1992b)Seung, Opper, and Sompolinsky]{SeungQBC}
Seung, H.~S., Opper, M., and Sompolinsky, H. (1992b).
\newblock Query by committee.
\newblock In {\em Proceedings of the Fifth Annual Workshop on Computational Learning Theory\/}, COLT '92, page 287–294, New York, NY, USA. Association for Computing Machinery.

\bibitem[Shannon(1948)Shannon]{shannon}
Shannon, C.~E. (1948).
\newblock A mathematical theory of communication.
\newblock {\em The Bell System Technical Journal\/}, {\bf 27}, 379--423.

\bibitem[Smith {\em et~al.}(2018)Smith, Nebgen, Lubbers, Isayev, and Roitberg]{smith2018less}
Smith, J.~S., Nebgen, B., Lubbers, N., Isayev, O., and Roitberg, A.~E. (2018).
\newblock Less is more: Sampling chemical space with active learning.
\newblock {\em The Journal of Chemical Physics\/}, {\bf 148}(24), 241733.

\bibitem[Street {\em et~al.}(1993)Street, Wolberg, and Mangasarian]{BreastCancerDataset}
Street, W.~N., Wolberg, W.~H., and Mangasarian, O.~L. (1993).
\newblock Nuclear feature extraction for breast tumor diagnosis.
\newblock In {\em Electronic imaging\/}.

\bibitem[Sutton and Barto(2018)Sutton and Barto]{sutton2018reinforcement}
Sutton, R.~S. and Barto, A.~G. (2018).
\newblock {\em Reinforcement learning: An introduction\/}.
\newblock MIT press.

\bibitem[Tang {\em et~al.}(2019)Tang, Li, and Huang]{alipy}
Tang, Y.-P., Li, G.-X., and Huang, S.-J. (2019).
\newblock {ALiPy}: Active learning in python.
\newblock Technical report, Nanjing University of Aeronautics and Astronautics.
\newblock available as arXiv preprint \url{https://arxiv.org/abs/1901.03802}.

\bibitem[Tietz {\em et~al.}(2017)Tietz, Fan, Nouri, Bossan, and {skorch Developers}]{skorch}
Tietz, M., Fan, T.~J., Nouri, D., Bossan, B., and {skorch Developers} (2017).
\newblock {\em skorch: A scikit-learn compatible neural network library that wraps PyTorch\/}.

\bibitem[Tsanas and Xifara(2012)Tsanas and Xifara]{energydataset}
Tsanas, A. and Xifara, A. (2012).
\newblock Accurate quantitative estimation of energy performance of residential buildings using statistical machine learning tools.
\newblock {\em Energy and Buildings\/}, {\bf 49}, 560--567.

\bibitem[Tüfekci(2014)Tüfekci]{powerdataset}
Tüfekci, P. (2014).
\newblock Prediction of full load electrical power output of a base load operated combined cycle power plant using machine learning methods.
\newblock {\em International Journal of Electrical Power \& Energy Systems\/}, {\bf 60}, 126--140.

\bibitem[Wang {\em et~al.}(2019)Wang, Pruksachatkun, Nangia, Singh, Michael, Hill, Levy, and Bowman]{wang2019superglue}
Wang, A., Pruksachatkun, Y., Nangia, N., Singh, A., Michael, J., Hill, F., Levy, O., and Bowman, S. (2019).
\newblock Superglue: A stickier benchmark for general-purpose language understanding systems.
\newblock {\em Advances in Neural Information Processing Systems\/}, {\bf 32}.

\bibitem[Wu(2019)Wu]{dongruiWURegressionAL}
Wu, D. (2019).
\newblock Pool-based sequential active learning for regression.
\newblock {\em IEEE Transactions on Neural Networks and Learning Systems\/}, {\bf 30}(5), 1348--1359.

\bibitem[Xiao {\em et~al.}(2017)Xiao, Rasul, and Vollgraf]{xiao2017/online}
Xiao, H., Rasul, K., and Vollgraf, R. (2017).
\newblock Fashion-mnist: a novel image dataset for benchmarking machine learning algorithms.

\bibitem[Xiong {\em et~al.}(2014)Xiong, Azimi, and Fern]{xiongAL}
Xiong, S., Azimi, J., and Fern, X.~Z. (2014).
\newblock Active learning of constraints for semi-supervised clustering.
\newblock {\em IEEE Transactions on Knowledge and Data Engineering\/}, {\bf 26}(1), 43--54.

\bibitem[Yang and Loog(2018)Yang and Loog]{yangloog}
Yang, Y. and Loog, M. (2018).
\newblock A benchmark and comparison of active learning for logistic regression.
\newblock {\em Pattern Recognition\/}, {\bf 83}, 401--415.

\bibitem[Yang {\em et~al.}(2017)Yang, Lee, Chung, Wu, Chen, and Lin]{libact}
Yang, Y.-Y., Lee, S.-C., Chung, Y.-A., Wu, T.-E., Chen, S.-A., and Lin, H.-T. (2017).
\newblock libact: Pool-based active learning in python.
\newblock Technical report, National Taiwan University.
\newblock available as arXiv preprint \url{https://arxiv.org/abs/1710.00379}.

\bibitem[Yeh(1998)Yeh]{concretedataset}
Yeh, I.-C. (1998).
\newblock Modeling of strength of high-performance concrete using artificial neural networks.
\newblock {\em Cement and Concrete Research\/}, {\bf 28}(12), 1797--1808.

\bibitem[Zagidullin {\em et~al.}(2019)Zagidullin, Aldahdooh, Zheng, Wang, Wang, Saad, Malyutina, Jafari, Tanoli, Pessia, and Tang]{drugcombdb}
Zagidullin, B., Aldahdooh, J., Zheng, S., Wang, W., Wang, Y., Saad, J., Malyutina, A., Jafari, M., Tanoli, Z., Pessia, A., and Tang, J. (2019).
\newblock {DrugComb: an integrative cancer drug combination data portal}.
\newblock {\em Nucleic Acids Research\/}, {\bf 47}(W1), W43--W51.

\bibitem[Zhan {\em et~al.}(2021)Zhan, Liu, Li, and Chan]{zhanreview}
Zhan, X., Liu, H., Li, Q., and Chan, A.~B. (2021).
\newblock A comparative survey: Benchmarking for pool-based active learning.
\newblock In Z.-H. Zhou, editor, {\em Proceedings of the Thirtieth International Joint Conference on Artificial Intelligence, {IJCAI-21}\/}, pages 4679--4686. International Joint Conferences on Artificial Intelligence Organization.
\newblock Survey Track.

\end{thebibliography}
\bibliographystyle{natbib}

\appendix
\section{Active learning primer}\label{apppendix:primer}

\paragraph{Notations} Let $\mathcal{X}$ denote a state space whereby each element $x\in\mathcal{X}$ can be associated to a label $y\in\mathcal{Y}$. We further denote by $\mathcal{F}$ the space of functions mapping $f: \mathcal{X}\mapsto\mathcal{Y}$. For a given ML architecture, we denote by $\mathcal{F}_\Theta\subset\mathcal{F}$ the subset of function mappings having this architecture and parameters in $\Theta$, which is defined by the model's hyperparameters. The \textit{world},  $\mathcal{W}\subset\mathcal{X}\times\mathcal{Y}$, is defined by a canonical mapping $\omega \in \mathcal{F}$ that associates to each state a unique label. We assume we have initially access to a set of observations $\mathcal{L}_0\subset\mathcal{W}$ from the world. We denote by $\mathcal{U}_0\subset\mathcal{X}$ the set of elements that are not observed in $\mathcal{L}_0$. The task we are faced with is to approximate the canonical mapping function $\omega$ from a subset of observations of $\mathcal{W}$, i.e. find $f\in\mathcal{F}$ such that $\forall x\in\mathcal{X}, f(x) \approx \omega(x)$.

\paragraph{Supervised learning} To address this task in the supervised ML paradigm, one first defines a function space $\mathcal{F}_{\Theta}$, through various inductive biases and hyperparameter tuning. Then one typically searches for a function $f_{\hat{\theta}} \in \mathcal{F}_{\Theta}$, with parameters $\hat{\theta} \in \Theta$, that best approximate $\omega$.

\paragraph{Active learning} In contrast, in the AL paradigm the function space $\mathcal{F}_{\Theta}$ is largely assumed to be fixed, but the practitioner can iteratively increase the size of the dataset by strategically querying the labels for some unlabelled elements in $\mathcal{X}$. At the $k^{\textnormal{th}}$ iteration, we denote by $\mathcal{L}_k$ the dataset and by $\mathcal{U}_k$ the set of unlabelled elements. The goal of an AL strategy is to identify a query set $\mathcal{Q}_k\subset\mathcal{U}_k$ that carries the most \textit{information} for approximating $\omega$. In other words, the AL practitioner looks to add new observations strategically to identify a smaller space $\mathcal{F}_{\Theta,k+1}$ of functions that are compatible with $\mathcal{L}_{k+1} =  \mathcal{L}_k\bigcup\big\{\left(q,\omega(q)\right),\, q\in\mathcal{Q}_k\big\}$, whereby the queries' labels are provided by an \textit{oracle}, e.g. human annotators or laboratory experiments. We then update $\mathcal{U}_{k+1} = \mathcal{U}_k\setminus\mathcal{Q}_k $ for the next iteration. Access to the oracle is costly, thus AL typically operates under a \textit{budget constraint} limiting the total number of possible queries.

The general framing described here is sometimes referred to as pool-based AL \cite{burrsettlesbook}, where $\mathcal{U}_k$ defines the \textit{pool} at the $k^{\textnormal{th}}$ iteration. This is arguably the most common framing of the AL problem. However, there exist variations whereby at each iteration $\mathcal{U}_k$ is not a finite set but either a generative process or a stream of unlabelled observation. These are known as membership query and stream based AL, respectively \cite{burrsettlesbook}. All framings share the same core problem: identifying unlabelled observations that are most \textit{informative} to improving our solution to the task. In our exposition, we will principally focus on pool-based AL unless otherwise specified.

A critical question underpinning AL is how to measure the \textit{informativeness} of each state in $\mathcal{X}$. A number of heuristics have been proposed in the past which can broadly be split into two categories: (1) uncertainty-based category and (2) diversity-based category. These are not mutually exclusive as the two notions can be combined to positive effect \cite{ burrsettlesbook, brinkerdiversityanduncertainty, monarch_human---loop_2021} and this is the view we take. The former relies on some measure of uncertainties of the predictions obtained from the set of functions $\mathcal{F}_{\Theta,k}$ at each iteration. In practice, $\mathcal{F}_{\Theta,k}$ can be defined intrinsically by the class of ML models (e.g. Gaussian processes or Bayesian neural networks) or a subset of it can be obtained using heuristics such as ensembling \cite{lakshminarayanan_simple_2017, Cohn1994, SeungQBC} or Bayesian inference approximation methods like MCDropout \cite{gal2016dropout}. Various informativeness scores can be computed from the uncertainty estimates and used to select the query set $\mathcal{Q}_k$. In contrast, diversity-based heuristics rely on a distance measure $d: \mathcal{X}\times\mathcal{X}\mapsto\mathbb{R}_+$, which can depend on $\mathcal{F}_{\Theta, k}$, to score states in $\mathcal{U}_k$ with respect to elements in $\mathcal{L}_k$. The choice of informativeness measure is an important challenge of AL as well as the choice of query selection strategy from the informativeness scores.

A standard approach to select $\mathcal{Q}_k$ is to greedily pick the highest scoring elements such that \[Q_k =  \argmax_{\substack{\mathcal{Q}\subset\mathcal{U}_k \\  |\mathcal{Q}|=m_k }} \sum_{q\in\mathcal{Q}}  I(q),\] where $I: \mathcal{X}\mapsto \mathbb{R}$ denotes an information measure that typically would depend on $\mathcal{F}_{\Theta,k}$, and $m_k$ corresponds to the number of queries desired at round $k$. However, for $m_k>1$, this selection strategy can lead to a redundant query set as each query therein is selected independently of the others.  The sub-field of \textit{batch AL} is concerned with this issue and developing strategies to ensure some notion of diversity in the query \cite{burrsettlesbook, brinkerdiversityanduncertainty, guobatchmode}. For instance, BatchBALD \cite{kirsch2019batchbald} relies on minimising the mutual information between pairs of queries in $\mathcal{Q}_k$.

\section{Informativeness measures}\label{appendix:infos}

In this section, we provide more details about each type of informativeness measure discussed in the main and described some of the specific scores mentioned in Table \ref{tab:informativeness_measures}. 

\subsection{Uncertainty-based informativeness scores}

As discussed in Appendix \ref{apppendix:primer}, a large number of informativeness measures rely on some estimate of prediction uncertainties. The underlying rationale is that regions of the world are highly uncertain because they are poorly represented in the training data. Improving coverage through sampling uncertain data regions can enable significant global performance gains. While this makes intuitive sense, it ignores subtleties about the nature of uncertainties which can be decomposed in epistemic and aleatoric components \cite{der2009aleatory}. The former is understood as coming from modelling errors while the latter is inherent to the world, corresponding to natural variations or measurement noise. Aleatoric uncertainties are irreducible and, in practice, it is extremely difficult to tease out the two uncertainty components from each other. This means that uncertainty-based informativeness scores can overestimate the information gain that a sample would provide. 

In the following, we detail a few of the scores that fall under this category. 

\paragraph{Entropy} is a well-established notion in information theory that quantifies the uncertainty of a probability distribution over possible outcomes \cite{shannon}. It is used as an informativeness measure for classification tasks where the probability distribution is defined over the finite set of labels $\mathcal{Y}$. Specifically, it is defined as \[I(x) = -\sum_{y\in\mathcal{Y}}p_{\mathcal{F}_{\theta,k}}\left(y|x\right)\log\left(p_{\mathcal{F}_{\theta,k}}\left(y|x\right)\right).\]

\paragraph{Least confidence} can be used both for regression and classification tasks, although the score is defined differently for each task. In regression tasks, we have \[I(x) = \textnormal{Var}_{\mathcal{F}_{\theta,k}}\left[Y | x\right],\] where Y denotes a random variable in $\mathcal{Y}$. Hence, to each sample $x$ it associates the variance across the predictions from the set of functions $\mathcal{F}_{\theta,k}$.

In classification tasks, the least confidence score is defined as \[I(x) = 1 - \max_{y\in\mathcal{Y}}\, p_{\mathcal{F}_{\theta,k}}\left(y|x\right).\]

\paragraph{Margin confidence} was introduced for classification tasks and measures uncertainty as the difference between the highest and second highest probabilities associated to labels in $\mathcal{Y}$ derived from our predictions on a given sample. The smaller that difference, the higher the uncertainty. In equation, this gives

\begin{align*}
    y_0 &= \argmax_{y\in\mathcal{Y}} p_{\mathcal{F}_{\theta,k}}\left(y|x\right),\\
    y_1 &= \argmax_{y\in\mathcal{Y}\setminus\{y_0\}}p_{\mathcal{F}_{\theta,k}}\left(y|x\right),\\
    I(x) &= 1 - \left(p_{\mathcal{F}_{\theta,k}}\left(y_0|x\right)-p_{\mathcal{F}_{\theta,k}}\left(y_1|x\right)\right).
\end{align*}

\paragraph{Ratio of confidence} is a score similar to margin confidence, the only difference comes from the fact that the uncertainty is expressed as a ratio between the two probabilities
\begin{align*}
    y_0 &= \argmax_{y\in\mathcal{Y}} p_{\mathcal{F}_{\theta,k}}\left(y|x\right),\\
    y_1 &= \argmax_{y\in\mathcal{Y}\setminus\{y_0\}}p_{\mathcal{F}_{\theta,k}}\left(y|x\right),\\
    I(x) &= \frac{p_{\mathcal{F}_{\theta,k}}\left(y_1|x\right)}{p_{\mathcal{F}_{\theta,k}}\left(y_0|x\right)}.
\end{align*}

\paragraph{Average value score.} This is a simple score used for regression tasks where greater values in $\mathcal{Y}$ are of higher interest. For instance, we would be more interested in reliably identifying highly synergistic pairs of drugs. The score is defined as \[I(x) = \textnormal{E}_{\mathcal{F}_{\theta,k}}\left[Y | x\right].\]

\paragraph{Upper confidence bound} UCB comes from the reinforcement learning (RL) literature and corresponds to a trade-off, controlled by an hyperparameter $\lambda$, between the greedy score and  the least confidence score. It is formally defined as \[I(x) = \textnormal{E}_{\mathcal{F}_{\theta,k}}\left[Y|x\right] +\lambda \sqrt{\textnormal{Var}_{\mathcal{F}_{\theta,k}}\left[Y|x\right]},\]
where Y denotes a random variable in $\mathcal{Y}$.

Similarly to the greedy score, it is defined for continuous label values and assumes that greater values, known as rewards in RL, are more relevant to the problem.

\subsection{Diversity-based informativeness scores}\label{app:diversity_scores}

Rather than rely on some estimate of uncertainties, diversity-based scores aim to increase the diversity of the training set, querying samples that are far from the current labelled set according to a selected distance metric. Methods can differ on the choice of distance metric and on the space where distances are computed. The ambient space is often used but can have some limitations due to noisy or redundant features or suffer from the curse of dimensionality in high-dimensional space. As such, researchers have proposed to compute distances in lower dimensional spaces either obtained from dimensionality reduction techniques or from intermediary latent spaces derived from a trained model. We discuss below two generic types of approaches that are built-in PyRelationAL.

\paragraph{Relative distance sampling} computes the minimum distance from an unlabelled sample to the labelled samples \[I(x) = \min_{l\in\mathcal{L}} d\left(\phi(x),\phi(l)\right),\] where $\phi:\mathcal{X}\mapsto \mathbb{R}^m$ an embedding function projecting samples into a latent space and $d:\mathbb{R}^m\times\mathbb{R}^m\mapsto \mathbb{R}_+$ represents the chosen distance metric in the latent space. Note that $\phi$ is simply the identity function when computing distances in the ambient space. The coreset approach, as defined in \cite{mehrjou_genedisco_2021}, would be classified in this category in PyRelationAL, where the latent space representation of a point is defined by the associated penultimate activations in a neural network.

\paragraph{Representative sampling} does not explicitly define an informativeness scores. Instead, it groups the current unlabelled set into a fixed number of clusters and selects representative samples from each cluster to form the query set. Note that this can be quite expensive to compute when the unlabelled set is large as it requires all-to-all pairwise distances, and can benefit from extracting first a subset of the unlabelled set based on some other criteria.

\section{Example model definition} \label{app:cnn_model}

To keep the main article light and focused on pyrelational contribution, we provide below the code associated with the model used in Section \ref{sec:case_study}.

\begin{code}

\begin{minted}[linenos,fontsize=\scriptsize,xleftmargin=0.5cm,numbersep=3pt,frame=lines]{python}
import torch
from lightning.pytorch import LightningModule
from torch import Tensor
from torch import nn as nn
from torch.nn import functional as F
from torchmetrics.classification \
    import MulticlassAccuracy

from pyrelational.model_managers \
    import LightningMCDropoutModelManager


class ConvNet(LightningModule):
    """Simple ConvNet for MNIST classification."""

    def __init__(self, num_classes: int):
        super().__init__()

        self.conv1 = nn.Conv2d(1, 32, kernel_size=5)
        self.conv1_drop = nn.Dropout2d()
        self.conv2 = nn.Conv2d(32, 64, kernel_size=5)
        self.conv2_drop = nn.Dropout2d()
        self.fc1 = nn.Linear(1024, 128)
        self.fc1_drop = nn.Dropout()
        self.fc2 = nn.Linear(128, num_classes)

        self.test_accuracy = MulticlassAccuracy(
            num_classes=num_classes
        )

    def forward(self, x: Tensor) -> Tensor:
        """Run forward pass of the model."""
        x = F.relu(
            F.max_pool2d(
                self.conv1_drop(self.conv1(x)), 
                2,
            )
        )
        x = F.relu(
            F.max_pool2d(
                self.conv2_drop(self.conv2(x)), 
                2,
            )
        )
        x = x.view(-1, 1024)
        x = F.relu(self.fc1_drop(self.fc1(x)))
        x = self.fc2(x)
        x = F.log_softmax(x, dim=1)
        return x

    def training_step(self, batch, batch_idx):
        """Run training step for the model."""
        x, y = batch
        logits = self(x)
        loss = F.nll_loss(logits, y)
        self.log("loss", loss)
        return loss

    def validation_step(self, batch, batch_idx):
        """Run validation step for the model."""
        x, y = batch
        logits = self(x)
        loss = F.nll_loss(logits, y)
        return loss

    def test_step(self, batch, batch_idx):
        """Run test step for the model."""
        x, y = batch
        logits = self(x)
        loss = F.nll_loss(logits, y)
        self.log("test_loss", loss)
        self.test_accuracy.update(logits, y)

    def on_test_epoch_end(self) -> None:
        acc = self.test_accuracy.compute()
        self.log("test_accuracy", acc)
        self.test_accuracy.reset()

    def configure_optimizers(self):
        """Configure optimizer for the model."""
        optimizer = torch.optim.Adam(
            self.parameters(),
            lr=0.001,
        )
        return optimizer

\end{minted}
\vspace{-1.5em}
\captionof{listing}{Simple CNN model for image classification.}\label{code:model_definition}

\end{code}

\section{Dataset summaries} \label{app:datasetsummaries}

\begin{itemize}
    \item \textbf{BreastCancer} \cite{BreastCancerDataset}: The BreastCancer dataset consists of 569 samples of breast cancer patients with 30 features. Features consist of radius, texture, perimeter, area, smoothness, compactness, concavity, concave points, symmetry and fractal dimension to predict the diagnosis of breast cancer (malignant or benign).
    \item \textbf{Checkerboard2x2} \cite{kseniaLAL}: The Checkerboard2x2 dataset is a synthetic classification dataset with 2,000 samples and 2 features and 1 output. The dataset is generated using a 2x2 checkerboard pattern with two classes.
    \item \textbf{Checkerboard4x4} \cite{kseniaLAL}: The Checkerboard4x4 dataset is a synthetic classification dataset with 2,000 samples and 2 features and 1 output. The dataset is generated using a 4x4 checkerboard pattern with two classes.
    \item \textbf{CreditCard} \cite{creditcard}: The CreditCard dataset consists of 284,807 samples of credit card transactions with 30 features. Features consist of Time, Amount and 28 anonymised features to predict fraudulent transactions. It is highly imbalanced with only 0.172\% of transactions being fraudulent posing a challenge for random sampling.
    \item \textbf{FashionMNIST} \cite{xiao2017/online}: The FashionMNIST dataset consists of 60,000 training images and 10,000 test images of fashion articles constructed by Zalando. The images are grayscale and have a resolution of 28x28 pixels. The dataset is often used as a replacement for the MNIST dataset due to its similar structure.
    \item \textbf{GaussianClouds} \cite{kseniaLAL}: The GaussianClouds dataset is a synthetic classification dataset with 11,000 samples and 2 features and 1 output. The dataset is generated using two Gaussian distributions representing two classes. The dataset is designed to be simple and easy to visualise active learning behaviour particularly at the intersection of the clouds.
    \item \textbf{Glass} \cite{UCI}: The Glass dataset consists of 214 samples of glass with 10 features. Features consist of Refractive Index, Sodium, Magnesium, Aluminium, Silicon, Potassium, Calcium, Barium and Iron to predict the type of glass (7 types).
    \item \textbf{MNIST} \cite{LeCunMNIST}: The MNIST dataset consists of 60,000 training images and 10,000 test images of handwritten digits. The images are grayscale and have a resolution of 28x28 pixels. The dataset is widely used for benchmarking machine learning algorithms.
    \item \textbf{Parkinsons} \cite{ParkinsonsDataset}: The Parkinsons dataset consists of 195 samples of Parkinson's disease patients with 22 features. Features consist of demographic information, medical history, and motor and non-motor symptoms to predict the severity of Parkinson's disease.
    \item \textbf{Striatum} \cite{ogstriatumsource, kseniaLAL}: The Striatum dataset consists of 20,000 samples of striatum neurons described by 272 features obtained from 3D Electron Microscopy stack of rat neural tissue. The task is to predict the presence of mitochondria as set out in Konyushkova et al. \cite{kseniaLAL}.
    \item \textbf{SynthClass1} \cite{kseniaLAL}: The SynthClass1 dataset is a synthetic classification dataset with 500 samples and 2 features and 1 output. The dataset is generated using two Gaussian distributions representing two classes. The dataset is designed to be simple and easy to visualise active learning behaviour.
    \item \textbf{SynthClass2} \cite{kseniaLAL}: The SynthClass2 dataset is a synthetic classification dataset with 500 samples and 2 features and 1 output. The dataset is designed to be more complex than SynthClass1 generated from six Gaussian blobs divided into two classes with varying degrees of overlap in space.
    \item \textbf{Airfoil} \cite{UCI}: The Airfoil Self Noise dataset consists of 1,503 samples of airfoil self-noise with 5 features. Features consist of Frequency, Angle of Attack, Chord Length, Free-Stream Velocity and Suction Side Displacement Thickness to predict the scaled sound pressure level.
    \item \textbf{Concrete} \cite{concretedataset}: The Concrete dataset consists of 1,030 samples of concrete with 8 features. Features consist of Cement, Blast Furnace Slag, Fly Ash, Water, Superplasticizer, Coarse Aggregate, Fine Aggregate and Age to predict the compressive strength of the concrete.
    \item \textbf{Diabetes} \cite{diabetesEfron}: The Diabetes dataset consists of 442 samples of diabetes patients with 10 features consisting of age, sex, body mass index, average blood pressure, and six blood serum measurements. The task is to predict the progression of diabetes one year after baseline.
    \item \textbf{DrugComb} \cite{drugcombdb}: The DrugComb dataset consists of 739,964 samples of drug combinations with drug-drug-dose inputs. Inputs consist of Drug1, Drug2 and Dose to predict the synergy score of the drug combination. In our library, we use the synergy score as the output for regression tasks with morgan fingerprints as features for each drug computed through RDKit \cite{rdkit}.
    \item \textbf{Energy} \cite{energydataset}: The Building Energy Efficiency Prediction dataset consists of 768 samples of building energy efficiency with 8 features. Features consist of Relative Compactness, Surface Area, Wall Area, Roof Area, Overall Height, Orientation, Glazing Area and Glazing Area Distribution to predict the heating load of the building. 
    \item \textbf{Power} \cite{powerdataset}: The Combined Cycle Power Plant dataset consists of 9,568 samples of electrical power output with 4 features. Features consist of hourly average ambient variables Temperature, Ambient Pressure, Relative Humidity and Exhaust Vacuum to predict the net hourly electrical energy output of the plant. 
    \item \textbf{SynthReg1 (Proposed)}: The SynthReg1 dataset is a synthetic regression dataset with 1,000 samples and 1 feature and 1 output. The dataset is generated using a random linear model with normally distributed noise. The dataset is designed to be simple and easy to visualise active learning behaviour.
    \item \textbf{SynthReg2 (Proposed)}: The SynthReg2 dataset is a synthetic regression dataset with 1,000 samples and 2 features and 1 output describing a spiral in 3 dimensions. The dataset is generated using periodic functions with normally distributed noise. The dataset is designed to be simple and easy to visualise active learning behaviour, under a more challenging scenario than SynthReg1.
    \item \textbf{WineQuality} \cite{winequalitydataset}: The WineQuality dataset consists of 1,598 samples of red and white wine with 11 physicochemical features and a quality score. 
    \item \textbf{Yacht} \cite{UCI}: The Yacht dataset consists of 306 samples of a yacht with 6 features by the Delft Ship Hydromechanics Laboratory to predict hydrodynamic performance.
\end{itemize}

\end{document}